\documentclass[letterpaper, 10 pt, conference]{ieeeconf}  % Comment this line out if you need a4paper

\usepackage{multirow}
\usepackage{lipsum} 
\usepackage{amssymb}
\usepackage{placeins}
\usepackage{amsmath} 
\usepackage{nicefrac}
\usepackage{graphicx}
\usepackage{color}
\usepackage{booktabs}
\usepackage{array}
\usepackage{bigdelim}
\usepackage{makecell} 
\usepackage{subcaption}
\usepackage[style=ieee]{biblatex}

\DeclareSourcemap{
  \maps{
    \map{
      \pertype{misc}
      \step[fieldset=language, null]
      \step[fieldset=url, null]
      \step[fieldset=doi, null]
      \step[fieldset=issn, null]
      \step[fieldset=isbn, null]
      \step[fieldset=editor, null]
      \step[fieldset=urldate, null]
      \step[fieldset=file, null]
    }
  }
}
\DeclareSourcemap{
  \maps{
    \map{
      \pertype{article}
      \step[fieldset=language, null]
      \step[fieldset=url, null]
      \step[fieldset=doi, null]
      \step[fieldset=issn, null]
      \step[fieldset=isbn, null]
      \step[fieldset=note, null]
      \step[fieldset=editor, null]
      \step[fieldset=urldate, null]
      \step[fieldset=file, null]
    }
  }
}
\DeclareSourcemap{
  \maps{
    \map{
      \pertype{inproceedings}
      \step[fieldset=language, null]
      \step[fieldset=url, null]
      \step[fieldset=doi, null]
      \step[fieldset=issn, null]
      \step[fieldset=isbn, null]
      \step[fieldset=note, null]
      \step[fieldset=editor, null]
      \step[fieldset=urldate, null]
      \step[fieldset=file, null]
    }
  }
}
\DeclareSourcemap{
  \maps{
    \map{
      \pertype{incollection}
      \step[fieldset=language, null]
      \step[fieldset=url, null]
      \step[fieldset=doi, null]
      \step[fieldset=issn, null]
      \step[fieldset=isbn, null]
      \step[fieldset=note, null]
      \step[fieldset=editor, null]
      \step[fieldset=urldate, null]
      \step[fieldset=file, null]
    }
  }
}

\IEEEoverridecommandlockouts                              % This command is only needed if 
\title{\LARGE \bf
Guiding Energy-Efficient Locomotion \\ through Impact Mitigation Rewards
}
\author{Chenghao Wang$^{1}$, Arjun Viswanathan$^{1}$, Eric Sihite$^{2}$, Alireza Ramezani$^{1*}$
\thanks{$^{1}$ The author is with Department of Electrical and Computer Engineering, Northeastern University, Boston, MA, USA  { wang.chengh, viswanathan.ar, a.ramezani@northeastern.edu}}%
\thanks{$^{2}$ The author is with the Department of Aerospace Engineering, California Institute of Technology, Pasadena, CA, USA { esihite@caltech.edu}}%
\thanks{$^{*}$ Corresponding author. Email: {a.ramezani@northeastern.edu}}%
}

\begin{document}

\maketitle
\thispagestyle{empty}
\pagestyle{empty}

%%%%%%%%%%%%%%%%%%%%%%%%%%%%%%%%%%%%%%%%%%%%%%%%%%%%%%%%%%%%%%%%%%%%%%%%%%%%%%%%
\begin{abstract}
Animals achieve energy-efficient locomotion by their implicit passive dynamics, a marvel that has captivated roboticists for decades. Recently, methods incorporated Adversarial Motion Prior (AMP) and Reinforcement learning (RL) shows promising progress to replicate Animals' naturalistic motion. However, such imitation learning approaches predominantly capture explicit kinematic patterns, so-called gaits, while overlooking the implicit passive dynamics. This work bridges this gap by incorporating a reward term guided by Impact Mitigation Factor (IMF), a physics-informed metric that quantifies a robot’s ability to passively mitigate impacts. By integrating IMF with AMP, our approach enables RL policies to learn both explicit motion trajectories from animal reference motion and the implicit passive dynamic. We demonstrate energy efficiency improvements of up to 32\%, as measured by the Cost of Transport (CoT), across both AMP and handcrafted reward structure.

\end{abstract}

%%%%%%%%%%%%%%%%%%%%%%%%%%%%%%%%%%%%%%%%%%%%%%%%%%%%%%%%%%%%%%%%%%%%%%%%%%%%%%%%
\section{INTRODUCTION}

Mimicry of bio-level energy-efficient locomotion has been a lasting aspiration in academia for decades \cite{collins_three-dimensional_2001}. Recent advancements in deep reinforcement learning have made this goal more attainable by allowing policies to explore within the action space to maximize specified reward functions. However, creating reward functions that effectively elicit the desired agent behaviors still demands significant specialized expertise and involves tedious engineering. On the other hand, imitation learning techniques enable agents to replicate reference motions by maximizing rewards for motion synthesis, either through complex motion tracking objectives or adversarial training with discriminators like Adversarial Motion Priors (AMP) \cite{peng_amp_2021}. While AMP simplifies reward engineering by distilling kinematic styles from motion capture (MoCap) data, it fundamentally overlooks the implicit passive dynamics, such as energy efficient impact absorption and compliance, that underpin biological locomotion.

Traditional handcrafted rewards, despite their success in enforcing task-specific stability through rewards (e.g., torque, velocity tracking), require laborious tuning to approximate energy-efficient gaits that mirror biological dynamics. This work bridges the gap between kinematic imitation and dynamic resilience by implementing the Impact Mitigation Factor \cite{wensing_proprioceptive_2017} (IMF) , a physics-informed metric that quantifies a robot’s passive ability to mitigate impacts based on its configuration. Thus empowers policies to learn not only explicit motion trajectories from MoCap data (via AMP) but also the implicit dynamic principles of biological systems. When combined with handcrafted rewards, the IMF enhances their inherent energy efficiency.

\begin{figure}[t!]
    \centering
    \includegraphics[width = 0.9\linewidth]{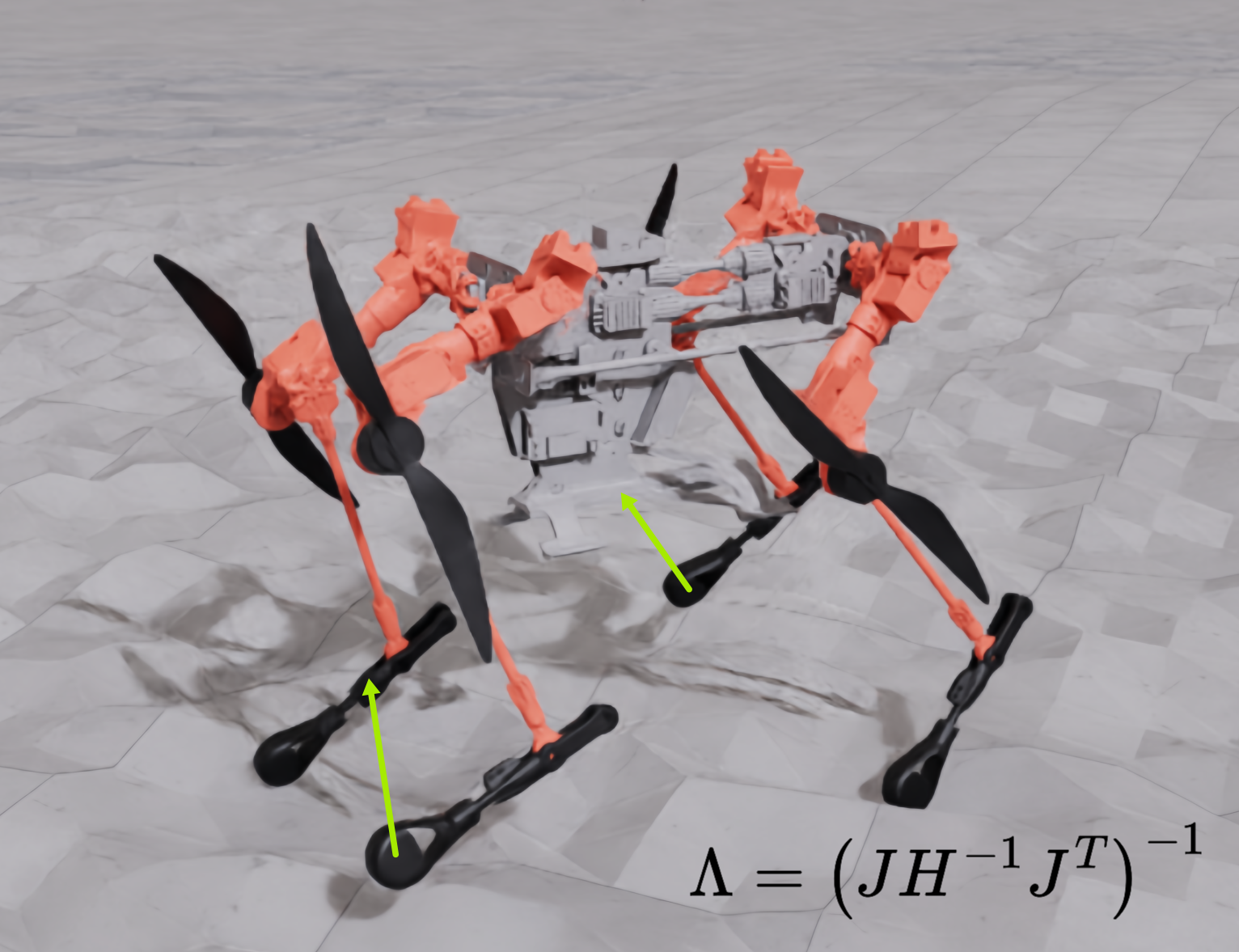}
    \caption{Arrows from the swing leg illustrate the pre-impact velocity and resulting impulse, both governed by the configuration-dependent Operational-Space Inertia Matrix ($\Lambda$). This matrix captures how the robot’s joint configuration influences impact dynamics, enabling effective mitigation.}
    % \vspace{-0.5cm}
    \label{fig:cover_imf}
\end{figure}

The primary contributions of this work are:
\begin{itemize}
    \item IMF as a Physics-Informed Reward Signal: Integrating the Impact Mitigation Factor (IMF) as a reward component that explicitly encodes passive compliance and impact resilience.
    \item Hybrid Reward Framework: Unifying AMP’s kinematic fidelity with the physics-informed IMF to achieve energy-efficient locomotion that captures both motion style and biomechanical passive dynamics.
    \item Experimental Validation: Demonstrating that augmenting AMP and handcrafted rewards with the IMF improves Cost of Transport (CoT) by 18\%-32\%.
\end{itemize}

\section{RELATED WORKS}

Reinforcement learning (RL) techniques have revolutionized legged locomotion by enabling policies to discover control strategies through trial-and-error. A pivotal development for motion imitation use RL is the integration of Adversarial Motion Priors (AMP) \cite{escontrela_adversarial_2022, lerario_learning_2024, peng_amp_2021}, which employ adversarial imitation learning to distill naturalistic motion styles from reference data. Unlike traditional handcrafted rewards that rely on meticulous tuning of kinematic tracking terms (e.g., pose, velocity) and penalty terms (e.g., torque, slippage), AMP trains a discriminator to distinguish agent-generated motions from MoCap data, explicitly encoding the reference kinematic patterns. This approach has enabled legged robots to emulate agile animal gaits \cite{escontrela_adversarial_2022}, transition seamlessly between walking and flying \cite{lerario_learning_2024}, and generalize skills from unstructured motion datasets \cite{peng_amp_2021}. While AMP simplifies reward engineering by focusing on kinematic fidelity, it does not explicitly account for the passive dynamics inherent in biological systems, such as energy transfer by impact absorption, which are critical for dynamic resilience and efficiency. Complementary methods like constrained RL \cite{lee_evaluation_2023} and hybrid model-based approaches \cite{kim_learning_2024} further enhance robustness but retain a focus on task-specific kinematic or force objectives rather than intrinsic physical interaction principles.

The Impact Mitigation Factor (IMF), introduced by \cite{wensing_proprioceptive_2017}, quantifies a robot’s ability to passively mitigate impacts by analyzing its operational-space inertia matrix. It compares the effective inertia during free motion to a "locked-joint" type motion, with higher values indicating superior impact absorption. This metric has guided the design of high-torque-density actuators in MIT Cheetah, enabling animal-like resilience without explicit compliance. Recent work \cite{relano_generalization_2023} extended the IMF to fixed-base robots as the Generalized Impact Absorption Factor (GIAF), optimizing backdrivability for collaborative systems. While the IMF has advanced mechanical design, its potential as a learning signal to bridge kinematic imitation and dynamic efficiency remains unexplored. Our work integrates the IMF into reinforcement learning, combining AMP’s motion priors with physics-informed resilience to address limitations in existing reward structures.

\section{Hardware Overview}

\begin{figure}[!]
\vspace{0.5cm}
    \centering
    \includegraphics[width = 0.6\linewidth]{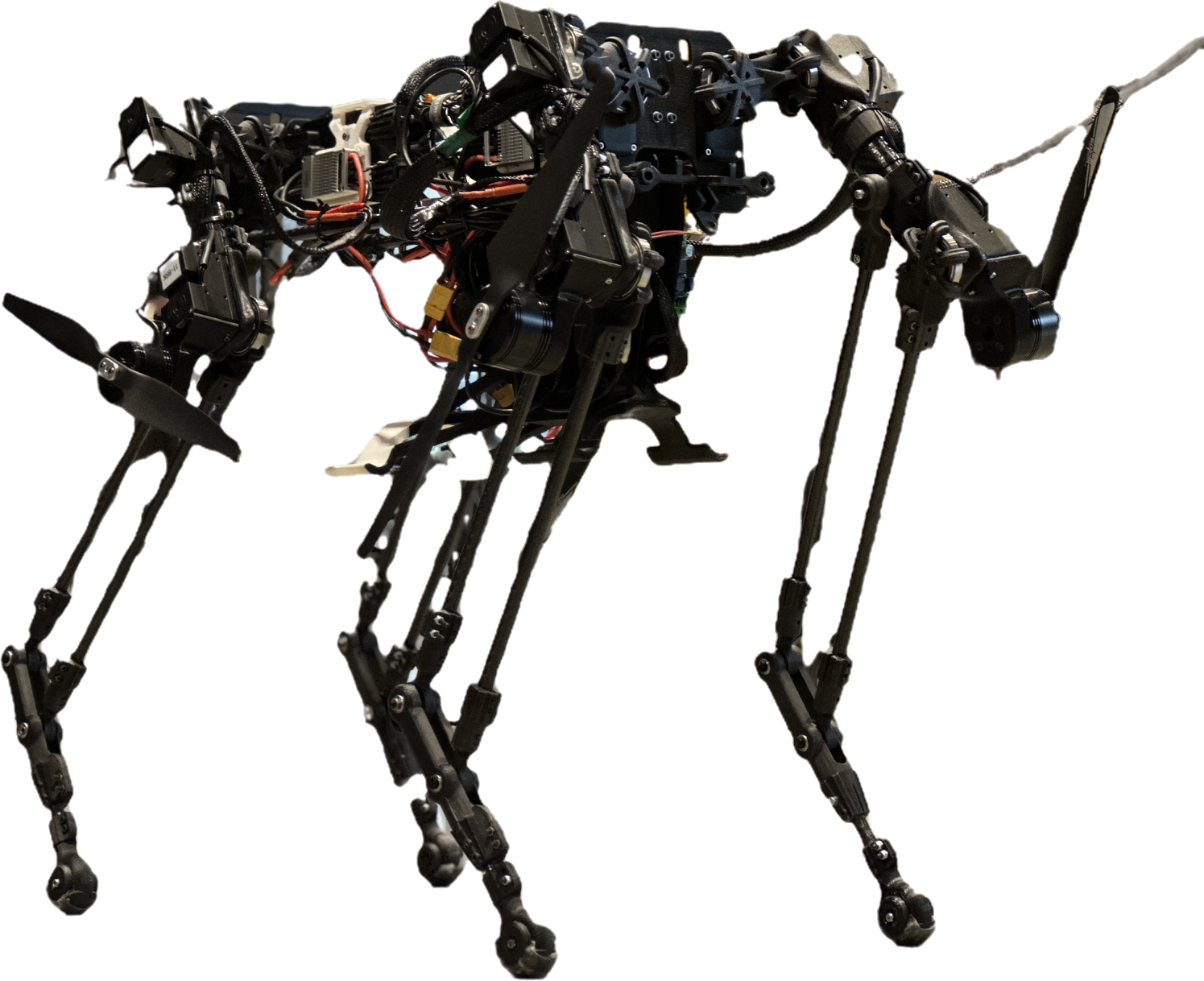}
    \caption{Northeastern University's Husky version-v.2 -- a platform designed to explore multi-modal dynamic-legged-aerial locomotion through appendage repurposing -- is the motivation for this study. }
    % \vspace{-0.5cm}
    \label{fig:hw-overview}
\end{figure}
Husky-v.2 \cite{dangol_performance_2020,de_oliveira_thruster-assisted_2020,salagame_quadrupedal_2023,dangol_towards_2020,salagame_letter_2022,sihite_efficient_2022,sihite_multi-modal_2023,ramezani_generative_2021,dangol_control_2021,sihite_unilateral_2021,sihite_optimization-free_2021,pitroda_capture_2024,krishnamurthy_enabling_2024-1,krishnamurthy_narrow-path_2024,krishnamurthy_optimization_2024-1}, shown in Fig.\ref{fig:hw-overview} is engineered for robust quadrupedal locomotion and aerial mobility through a bio-inspired morphing mechanism that enables seamless transitions between legged and UAV modes. The robot features four BLDC-actuated propellers mounted at its knee joints, allowing it to generate upward thrust by extending a leg and orienting the propellers upward, like a conventional quadcopter. The primary design rationale for such a hybrid robot is balancing actuation power with overall weight, which is critical for achieving stable flight without sacrificing ground mobility.

To address the challenge, our design emphasizes several strategies. A lightweight structure is achieved by using materials such as 3D-printed Onyx polymer and carbon fiber for the main body and leg assemblies, which minimizes the robot's mass. High torque-to-weight actuators were selected; for example, the Dynamixel XH540-W270-T servo provides a stall torque of 10.6~Nm while weighing only 165~g, offering strong leg actuation with minimal weight penalty. Additionally, each leg is equipped with a compact propulsion unit—a combination of a SunnySky X4112S brushless motor, an EOLO 50A LIGHT ESC, and a 15\(\times\)5.5 double-blade propeller—that collectively generates up to 12~kg of thrust, yielding a thrust-to-weight ratio of approximately 1.8 ensuring sufficient thrust for stable flight.

In this work, we focus exclusively on low-level locomotion policy learning. Accordingly, the robot's flight capability, while inherent to the design, is not demonstrated in in this work to maintain clarity of scope and avoid potential confusion.

\section{Impact mitigation factor (IMF)}
In this section, we outline the methodology for computing the Impact Mitigation Factor (IMF), denoted as \(\xi\), and its associated reward term, \(R_{\text{IMF}}\), for floating-base robotic systems. The IMF quantifies a robot's capacity to mitigate impact forces at its end-effector by comparing the impact response of the full dynamic system to that of a locked-joint scenario. The robot is expected to implicitly learn to convert impact energy into passive dynamics by adopting configurations that maximize \(R_{\text{IMF}}\).

\subsection{System Description}

Consider a floating-base robot with a configuration vector defined as:
\begin{equation}
q = \begin{bmatrix} q_b \\ q_j \end{bmatrix} \in \mathbb{R}^{n+6}
\end{equation}
where \( q_b \in \mathbb{R}^6 \) represents the position and orientation of the floating base, and \( q_j \in \mathbb{R}^n \) denotes the joint coordinates of the robot's \( n \) joints.

The system dynamics are governed by the following equation:
\begin{equation}
\begin{bmatrix} H_{bb} & H_{bj} \\ H_{jb} & H_{jj} \end{bmatrix} \begin{bmatrix} \ddot{q}_b \\ \ddot{q}_j \end{bmatrix} + h(q, \dot{q}) = S^T \tau + J^T f
\end{equation}\\
where \(H \in \mathbb{R}^{(n+6) \times (n+6)}\) is the inertia matrix partitioned into \(H_{bb}\), \(H_{bj}\), \(H_{jb}\), and \(H_{jj}\) for the base and joints; \(h(q, \dot{q}) \in \mathbb{R}^{n+6}\) represents Coriolis, centrifugal, and gravitational forces; \(\tau \in \mathbb{R}^n\) and \(f \in \mathbb{R}^m\) denote the joint torques and end-effector forces, respectively; \(J = \begin{bmatrix} J_b & J_j \end{bmatrix} \) is the Jacobian that maps generalized velocities to the end-effector velocity with \( J_b \in \mathbb{R}^{m \times 6} \) and \( J_j \in \mathbb{R}^{m \times n} \); and \(S \in \mathbb{R}^{n \times (n+6)}\) selects the actuated joints.
\subsection{Rigid-Body Impact Dynamics}

For impact analysis, we consider the velocity at the impact point (e.g., the end-effector) as:
\begin{equation}
v = J \dot{q}
\end{equation}
where \( v \in \mathbb{R}^m \) is the velocity that the system hits the ground with. Upon an impact causing a velocity change, the resulting contact impulse \( \rho \in \mathbb{R}^3 \) is:
\begin{equation}
\rho = -\Lambda v
\end{equation}
where \( \Lambda \) is the Operational-Space Inertia Matrix (OSIM), defined as:
\begin{equation}
\Lambda = (J H^{-1} J^T)^{-1}
\end{equation}
The matrix \( \Lambda \in \mathbb{R}^{m \times m} \) represents the effective inertia felt at the end-effector, accounting for the floating-base dynamics fully.

\subsection{Locked-Case}

When the reflected inertias at the joints become exceedingly high, the robot behaves as if its joints are completely locked. In this extreme case, only the base contributes to the motion, and the Jacobian effectively reduces to its base component \(J_b\). We then define the locked-case operational-space inertia as
\begin{equation}
\Lambda_L = \left( J_b\,H_{bb}^{-1}\,J_b^T \right)^{-1}
\end{equation}
where \(H_{bb} \in \mathbb{R}^{6 \times 6}\) is the inertia submatrix corresponding to the base. The impact impulse in this rigid, locked scenario is given by
\begin{equation}
\rho_L = -\Lambda_L\,v
\end{equation}
This locked-case inertia \(\Lambda_L\) provides a benchmark against which the benefits of joint compliance can be measured, and is typically larger than the effective inertia when joints are free to move.

\subsection{Impact Mitigation Factor (IMF)}

The IMF, \( \xi \), is derived by comparing the actual impact impulse \( \rho \) with the locked-case impulse \( \rho_L \). Substituting \( v = -\Lambda_L^{-1} \rho_L \) into the expression for \( \rho \), we obtain:
\begin{equation}
\rho = -\Lambda (-\Lambda_L^{-1} \rho_L) = \Lambda \Lambda_L^{-1} \rho_L
\end{equation}
The difference between the locked and actual impulses highlights the mitigation effect of the joints' free dynamics:
\begin{equation}
\rho_L - \rho = \rho_L - \Lambda \Lambda_L^{-1} \rho_L = (I - \Lambda \Lambda_L^{-1}) \rho_L
\end{equation}
where \( I \in \mathbb{R}^{m \times m} \) is the identity matrix. We define the impact mitigation matrix as:
\begin{equation}
\Xi = I - \Lambda \Lambda_L^{-1}
\end{equation}
The IMF is then a scalar measure computed as the determinant of \( \Xi \):
\begin{equation}
\xi = \det(\Xi)
\end{equation}
Since \( \Lambda \) and \( \Lambda_L \) are positive definite matrices, the eigenvalues of \( \Lambda \Lambda_L^{-1} \) are positive, and those of \( \Xi \) are between 0 and 1. Thus, \( \xi \) ranges from 0 to 1:
\begin{itemize}
    \item \( \xi = 1 \) indicates perfect impact absorption, where \( \rho \approx 0 \) (complete mitigation compared to the locked case),
    \item \( \xi = 0 \) indicates no mitigation, where \( \rho \approx \rho_L \) (the response matches the locked case).
\end{itemize}

\subsection{IMF as reward}

The IMF \( \xi \) is a configuration-dependent metric that quantifies the robot's ability to passively absorb impacts through its dynamic properties. Given that \( \xi \) ranges from 0 to 1, it is particularly well-suited for integration into reinforcement learning frameworks as a reward signal. Specifically, the reward term can be defined as:
\begin{equation}
R^{\mathrm{IMF}} = -\log(1 - \xi)
\end{equation}

When \( \xi \) approaches 0, we have \( 1 - \xi \approx 1 \) and therefore \(R_{IMF} = -\log(1) = 0\). Conversely, as \( \xi \) approaches 1, \( 1 - \xi \) tends toward 0, causing \(R_{IMF} \to -\log(0) \to +\infty\).  This indicates that maximizing \( R_{IMF} \) will incentivize the RL policy to adopt configurations that enhance the robot's passive impact absorption capability. In other words, by exploiting its passive dynamics, the robot can implicitly learn an energy-efficient gait.

\section{Reinforcement Learning Formulation}
In this section, we formulate the Reinforcement Learning (RL) problem as a Markov Decision Process (MDP). An MDP is defined by a set of states \(\mathcal{S}\), a set of actions \(\mathcal{A}\), a transition probability function \(P(s_{t+1} \mid s_t, a_t)\), a reward function \(R(s_t, a_t, s_{t+1})\), and a discount factor \(\gamma\). At each time step \(t\), the agent observes the current state \(s_t \in \mathcal{S}\) and selects an action \(a_t \in \mathcal{A}\). The environment then transitions to a new state \(s_{t+1}\) according to the probability \(P(s_{t+1} \mid s_t, a_t)\), and the agent receives a reward \(r_t = R(s_t, a_t, s_{t+1})\). The objective is to learn a policy \(\pi(a_t \mid s_t)\) that maximizes the expected discounted return \(\mathbb{E} \left[ \sum_{t=0}^{\infty} \gamma^t r_t \right]\). 

To evaluate the effectiveness of our proposed IMF-guided reward, we define four reward structures: two baselines-Adversarial Motion Priors (AMP) with a task reward and a hand-designed reward and their IMF-augmented versions. The following of this section detail these reward formulations and their implementation.

\subsection{Adversarial Motion Priors (AMP)-Based Reward}
AMP utilize adversarial learning to extract style rewards directly from reference motion datasets, hence reducing the dependence on tedious hand-crafted reward functions. When combined with task reward, this AMP-based style reward steers the policy toward achieving its objectives while following reference behavior. 

However, the AMP reward based policy mainly captures the explicit kinematic patterns in reference motion rather than implicit passive dynamics.  To address this limitation, we incorporate the IMF into our reward formulation, effectively compensating for the absent passive dynamics and enabling a more comprehensive imitation of natural motion.
\subsubsection{Reference data pre-process}
The motion reference data from \cite{zhang_mode-adaptive_2018} comprises clips of unstructured dog motion recorded on flat terrain, capturing a variety of locomotion modes—such as walking, pacing, and trotting—with a total duration of 4.5 seconds. Following the kinematic remapping method described in \cite{peng_learning_2020}, the target motion is retargeted from the source morphology to that of Husky-v. 2. The Pinocchio \cite{carpentier_pinocchio_2024} library is employed to compute both forward and inverse kinematics. By recording the positions and velocities of the robot base, joints, and leg end, and then by combining states from all locomotion phases, we obtain a reference motion dataset for generating the style reward in discriminator training.

\subsubsection{Discriminator for generating style reward}
To train the discriminator for generating the style reward in AMP, a neural network \(D_\phi\), parameterized by \(\phi\), is employed to distinguish between state transitions from a reference motion dataset \(\mathcal{D}\) and those produced by the policy \(\pi_\theta\). The reference dataset \(\mathcal{D}\) contains motion reference in the robot’s morphology, while policy transitions are generated from policy output actions during training. 

The discriminator is trained using a least squares Generative Adversarial Network (GAN) formulation \cite{peng_amp_2021}, aiming to output 1 for reference transitions \((s, s') \sim \mathcal{D}\) and -1 for policy-generated transitions \((s, s') \sim \pi_\theta\). Its training objective is to minimize the following loss function:

\begin{equation}
\begin{split}
\mathcal{L}_D =\ & \mathbb{E}_{(s,s') \sim \mathcal{D}} \left[ (D_\phi(s, s') - 1)^2 \right] \\
& + \mathbb{E}_{(s,s') \sim \pi_\theta} \left[ (D_\phi(s, s') + 1)^2 \right] \\
& + w_{\text{gp}} \mathbb{E}_{(s,s') \sim \mathcal{D}} \left[ \| \nabla_\phi D_\phi(s, s') \|^2 \right]
\end{split}
\end{equation}\\

\noindent Here, the first term penalizes deviations from 1 for reference data, the second term penalizes deviations from -1 for policy data, and the gradient penalty term, weighted by \(w_{\text{gp}}\), is applied to reference samples to enhance training stability by encouraging smooth gradients.

To transform this output into a suitable reward signal, the style reward is defined as follows \cite{escontrela_adversarial_2022}:

\begin{equation}
r_t^\mathrm{Style}(s_t, s_{t+1}) = \max \Biggl[0,\; 1 - 0.25 \Bigl(D_\phi(s_t, s_{t+1}) - 1\Bigr)^2\Biggr]
% \tag{14}
\end{equation}

If \(D_\phi(s_t, s_{t+1})\) is close 
to 1, indicating a real transition, the squared term shrinks, making 
the reward approach 1. Conversely, if \(D_\phi(s_t, s_{t+1})\) is near 
\(-1\), suggesting a fake transition—the squared term grows, pushing the reward 
toward 0. 
\subsubsection{Task reward}
While the style reward focuses on enforcing the robot to follow a specific kinematic pattern of the reference motion. To successfully achieve locomotion, a task reward, which directs the policy to follow given commands such as velocity, is also necessary. 

The desired command is defined as
$\mathbf{v}_t = [v_{x,t}, v_{y,t}, \omega_{z,t}]^\top,
$ where \(v_{x}\) and \(v_{y}\) denote the forward and lateral velocities in the robot's base frame, and \(\omega_t\) is the desired yaw rate. The task reward at time $t$ is defined as

\begin{equation}
% \tag{15}
\small
r_t^\mathrm{Task} = w_v \exp\bigl(-\,\|\hat{\mathbf{v}}_{xy,t} - \mathbf{v}_{xy,t}\|\bigr)
      + w_\omega \exp\bigl(-|\hat{\omega}_{z,t} - \omega_{z,t}|\bigr)
\end{equation}

\noindent where \(w_v\) and \(w_\omega\) are weights, \(\hat{\mathbf{v}}_{xy,t}\) represents the measured planar velocity, and \(\hat{\omega}_{z,t}\) represents the measured yaw rate.

Finally, the AMP-based reward structure that incorporates the IMF is defined as:
\begin{equation}
% \tag{16}
r_t = w_\mathrm{Task}\, r_t^\mathrm{Task} + w_\mathrm{Style}\, r_t^\mathrm{Style} + w_{\mathrm{IMF}}\, r_t^{\mathrm{IMF}}
\end{equation}

\subsection{Hand-crafted Reward}
While the AMP-based method provides reference motion for the policy through kinematic guidance, handcrafted rewards allow the agent to freely explore the action space without being constrained by any reference. To validate the effectiveness of the proposed IMF-guided reward, we also implemented a handcrafted reward baseline.

We define the hand-crafted rewards using a structure similar to \cite{escontrela_adversarial_2022}, with further details provided in table \ref{tab:reward-structure}.

\begin{table}[ht]
  \centering
  \renewcommand{\arraystretch}{1.2} 
  \resizebox{\linewidth}{!}{%
    \begin{tabular}{l l r}
      \toprule
      Term & Expression & Weight \\ \midrule
      Linear Vel. Reward & $\exp\!\Bigl(-\frac{\|\hat{\mathbf{v}}_{xy,t} - \mathbf{v}_{xy,t}\|^2}{\sigma^2}\Bigr)$ & $1.5$ \\
      Angular Vel. Reward & $\exp\!\Bigl(-\frac{(\hat{\omega}_z - \omega_z)^2}{\sigma^2}\Bigr)$ & $0.75$ \\
      Vertical Vel. Penalty & $\hat{v}_z^2$ & $-2.0$ \\
      Angular (Roll/Pitch) Penalty & $\lVert\hat{\omega}_{xy}\rVert^2$ & $-0.05$ \\
      Orientation Penalty & $\|\mathrm{proj}_{xy}(\mathbf{g}_b)\|$ & $-0.5$ \\
      Height Deviation Penalty & $(h - H_{\text{Desired}})^2$ & $-2.0$ \\
      Torque Effort Penalty & $\|\tau\|_2$ & $-1\times10^{-5}$ \\
      Acceleration Effort Penalty & $\|\ddot{q}\|_2$ & $-1\times10^{-7}$ \\
      Position Limits Violation & $\sum_{i} \mathrm{penalty}(q_i)$ & $-2$ \\
      Action Smoothness Penalty & $\|\mathbf{a}_t - \mathbf{a}_{t-1}\|_2$ & $-0.01$ \\
      Limb Contact Penalty & $\mathbf{1}\{\text{contact on undesired areas}\}$ & $-1.0$ \\
      Foot Air Time Reward & $\displaystyle \sum_{\text{feet}}\Bigl[(t_{\text{last air}} - \mathrm{thr}) \cdot \mathbf{1}_{\{\text{first contact}\}}\Bigr]$ & $0.01$ \\
      Stand-Still Joint Deviation & $\sum_{i}\left|q_i - q_i^{\text{default}}\right|$ & $-0.5$ \\
      \bottomrule
    \end{tabular}
  }
\caption{Hard-crafted reward structure}
 \label{tab:reward-structure}
\end{table}

\begin{table}[ht]
  \centering
  \renewcommand{\arraystretch}{0.9}
  \resizebox{\linewidth}{!}{%
    \begin{tabular}{llccc ccc}
      \toprule
      \multicolumn{2}{c}{} & \multicolumn{3}{c}{\textbf{Cost of Transport (CoT)}} & \multicolumn{3}{c}{\textbf{Velocity RMSE}} \\ 
      \cmidrule(lr){3-5} \cmidrule(lr){6-8}
      \textbf{Terrain} & \textbf{Policy} & \rotatebox{0}{1 m/s} & \rotatebox{0}{1.5 m/s} & \rotatebox{0}{2 m/s} & Pitch & Roll & Yaw \\ 
      \midrule
      \multirow{4}{*}{Flat Ground} 
          & AMP w/o IMF         & 1.63  & 1.22  & 1.06  & 0.67  & 0.651 & 0.09  \\
          & AMP w/ IMF          & \textbf{1.33} & \textbf{1.08} & \textbf{1.03} & \textbf{0.57} & \textbf{0.650} & \textbf{0.08} \\
          & Hand-crafted w/o IMF  & 1.08  & 0.89  & 0.83  & 0.42  & 0.42  & 0.10  \\
          & Hand-crafted w/ IMF   & \textbf{0.96} & \textbf{0.86} & \textbf{0.66} & \textbf{0.40} & \textbf{0.38} & \textbf{0.09} \\
      \midrule
      \multirow{4}{*}{Rough Terrain} 
          & AMP w/o IMF         & 1.49  & 1.24  & 1.323 & 1.48  & 0.85  & 0.23  \\
          & AMP w/ IMF          & \textbf{1.38} & \textbf{0.99} & \textbf{1.322} & \textbf{1.18} & \textbf{0.69} & \textbf{0.19} \\
          & Hand-crafted w/o IMF  & 1.37  & 0.88  & 0.56  & 0.74  & 0.68  & \textbf{0.19} \\
          & Hand-crafted w/ IMF   & \textbf{1.11} & \textbf{0.72} & \textbf{0.54} & \textbf{0.60} & \textbf{0.59} & 0.20  \\
      \bottomrule
    \end{tabular}%
  }
  \caption{Comparison of Cost of Transport (CoT) and Velocity RMSE for various policies on Flat Ground and Rough Terrain. Values in bold indicate the best (lowest) performance.}
  \label{c}
\end{table}

Then, we sum all components of the reward above with the IMF reward to compute the final reward:
\begin{equation}
    r_t = w_\mathrm{IMF}r^{\mathrm{IMF}}_t +\sum_iw_ir^i_t
\end{equation}

\subsection{Model Representation}

\subsubsection{Observation space}

The observation space \( \mathcal{O} \in \mathbb{R}^{56} \) we used for policy training is comprises the following components: the linear velocity \( v_b \in \mathbb{R}^3 \), angular velocity \( \omega_b \in \mathbb{R}^3 \) and the gravity vector \( \mathbf{g}_b \in \mathbb{R}^3 \) in body frame; the desired velocity commands \( c \in \mathbb{R}^3 \), the joint positions \( \theta \in \mathbb{R}^{16} \) and velocities \( \dot{\theta} \in \mathbb{R}^{16} \) of the 16 joints with 12 actuated and 4 underacted, and the previous actions \( a_{\text{prev}} \in \mathbb{R}^{12} \). These elements are concatenated to form \( \mathcal{O} = \begin{bmatrix} v_b^\top & \omega_b^\top & g_b^\top & c^\top & \theta^\top & \dot{\theta}^\top & a_{\text{prev}}^\top \end{bmatrix}^\top \), ensuring that the agent has a holistic view of the system's states and the intended task. For the AMP reward-guided policy, the observation space is similar but excludes the linear and angular velocities to leverage the policy focusing on more natural motion behaviors.

The state input for the discriminator is a pair of consecutive states \([s_t, s_{t+1}] \in \mathbb{R}^{102}\), where each state \(s_t \in \mathbb{R}^{51}\) comprises the following components: the joint positions \(\theta \in \mathbb{R}^{16}\) and joint velocities \(\dot{\theta} \in \mathbb{R}^{16}\) the leg end positions with repsected to robot body \(\mathbf{p}_{\mathrm{foot}} \in \mathbb{R}^{12}\), computed via forward kinematics; the linear velocity \(v_b \in \mathbb{R}^3\) and angular velocity \(\omega_b \in \mathbb{R}^3\) in the body frame; base height \(z_b \in \mathbb{R}\).

\subsubsection{Action space}
The action space $\mathcal{A} \in \mathbb{R}^{12}$ consists of the joint position commands for the 12 actuated joints.
% \begin{table}[h]
%   \centering
%   \resizebox{\linewidth}{!}{%
%     \begin{tabular}{l l r}
%       \toprule
%       Term & Definition & Weight \\ \midrule
%       Base linear z velocity  & $\hat{v}_z^2$  & -2.0  \\
%       Base angular xy velocity  & $\lVert\hat{\omega}_{xy}\rVert^2$  & -0.05  \\
%       Flat orientation  & $\lVert g_t^{xy} \rVert^2$ & -0.5  \\
%       Base height  & $(\hat p_z-p_z)^2$ & -2.0  \\
%       Joint torques  & $\lVert \tau \rVert^2$ & -1e-5  \\
%       Joint accelerations  & $\lVert \ddot{\theta} \rVert^2$ & -1e-7  \\
%       Joint position limits  & $\lVert g_t^{xy} \rVert$ & -2  \\
%       Action rate  & $\lVert a_t - a_{t-1} \rVert^2$ & -0.01  \\
%       Undesired contacts  & $(I_{c,ll}+I_{c,a}) \backslash I_{c,f}$ & -1.0  \\
%       Linear velocity tracking  & $\exp\!\Bigl({\frac{-\lVert \hat{v}_{xy}-v_{xy} \rVert}{\sigma}}\Bigr)$ & 1.5  \\
%       Angular velocity tracking  & $\exp\!\Bigl({\frac{-\lVert \hat{\omega}_{xy}-\omega_{xy} \rVert}{\sigma}}\Bigr)$ & 0.75  \\
%       Feet air time  & $\sum_{feet}[(t_{air}-t_{th})\cdot I_c]$ & 0.01  \\
%       \bottomrule
%     \end{tabular}%
%   }
%   \caption{Rewards and penalties for the hand-crafted section}
%   \label{tab:rewards-and-penalties}
% \end{table}

\begin{figure*}[!ht]
    \centering
    \begin{subfigure}{\linewidth}
        \centering
        \includegraphics[width=\linewidth]{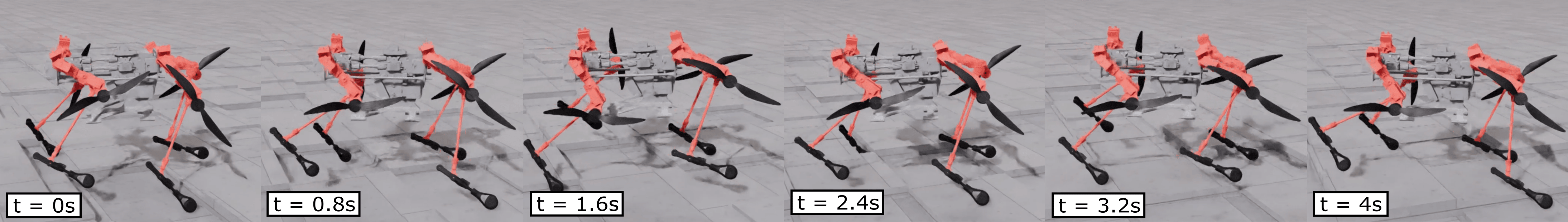}
        \caption{Snapshot of Husky v.2 policy utilizing hard-crafted reward without IMF.}
        \label{fig:snapshot-woIMF}
    \end{subfigure}

    \begin{subfigure}{\linewidth}
        \centering
        % \vspace{15pt}
        \includegraphics[width=\linewidth]{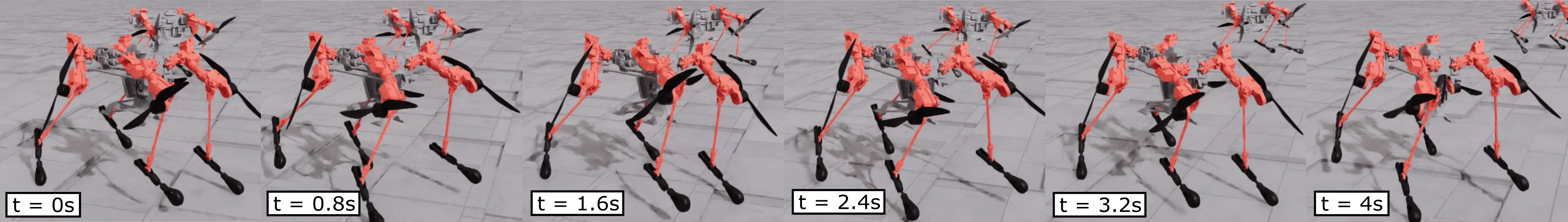} % update with your second figure's filename
        \caption{Snapshot of Husky v.2 policy utilizing hard-crafted reward with IMF.}
        \label{fig:snapshot-wIMF}
    \end{subfigure}

\caption{Snapshot of locomotion under policies trained with and without IMF. a): Rough terrain locomotion without IMF reward on Husky v.2. b): Rough terrain locomotion with IMF reward on Husky v.2. For the same trotting gait, the policy trained with IMF reward reduces average mechanical power consumption.}
    \label{fig:gait-comparsion}
    \end{figure*}

\section{Results}
\subsection{Experiment setup}
To evaluate the effectiveness of our proposed IMF-guided reward, we implemented four reward structures in the IsaacLab simulation environment: two baselines - one combining Adversarial Motion Priors (AMP) with a task reward and another using a hand-designed reward - and their corresponding IMF-augmented versions. For the IMF reward calculation, the robot’s simulated state is fed into Pinocchio to compute the necessary terms, such as the Jacobian and the inertia matrix etc. In the AMP-based reward structure, the weights were set at 2.0 for the AMP reward, 2.5 for the linear velocity tracking reward, and 1.5 for the angular velocity tracking reward. When the IMF reward was added, a weight of 10.0 was assigned to it. For the handcrafted reward combined with IMF, the IMF weight was 0.15 on flat ground and 1.5 on rough terrain.

The discriminator's neural network consists of two fully connected layers with 1024 and 512 neurons respectively (each followed by a LeakyReLU activation) and a final linear layer that produces a single scalar output. At the same time, the RL policy employs PPO where both the actor and critic networks include three hidden layers with 512, 256, and 128 neurons respectively, with ELU activations applied between layers.
\subsection{Simulation Results}
The experiment result demonstrate that augmenting AMP or handcrafted rewards with the IMF consistently improves energy efficiency and motion fidelity across terrains (Table \ref{tab:reward-structure}). On flat ground, the IMF-augmented AMP policy reduces the Cost of Transport (CoT, calculated as \(\nicefrac{\tau \cdot \dot{q}}{mgv}\)) by 18.4\% at 1 m/s and 2.8\% at 2 m/s compared to the baseline, while the handcrafted+IMF policy achieves a 20\% CoT reduction at 2 m/s. Notably, on rough terrain the handcrafted+IMF policy reduces CoT by 19.0\% at 1.5 m/s and maintains stable velocity tracking (Roll RMSE: 0.59 vs. 0.68 in the baseline). The AMP+IMF policy also shows significant improvements in dynamic resilience, with Pitch RMSE decreasing by 20.3\% on rough terrain.

% We evaluate the AMP approach with and without IMF and observe that incorporating IMF significantly reduces the cost of transport at moderately high commanded velocities. However, at higher velocities, the cost reduction becomes negligible. This is likely because, at such speeds, ground impact forces are already rapidly transmitted through the legs, limiting the additional benefit of IMF.
\begin{figure}[!ht]
    \centering
    \includegraphics[width = \linewidth]{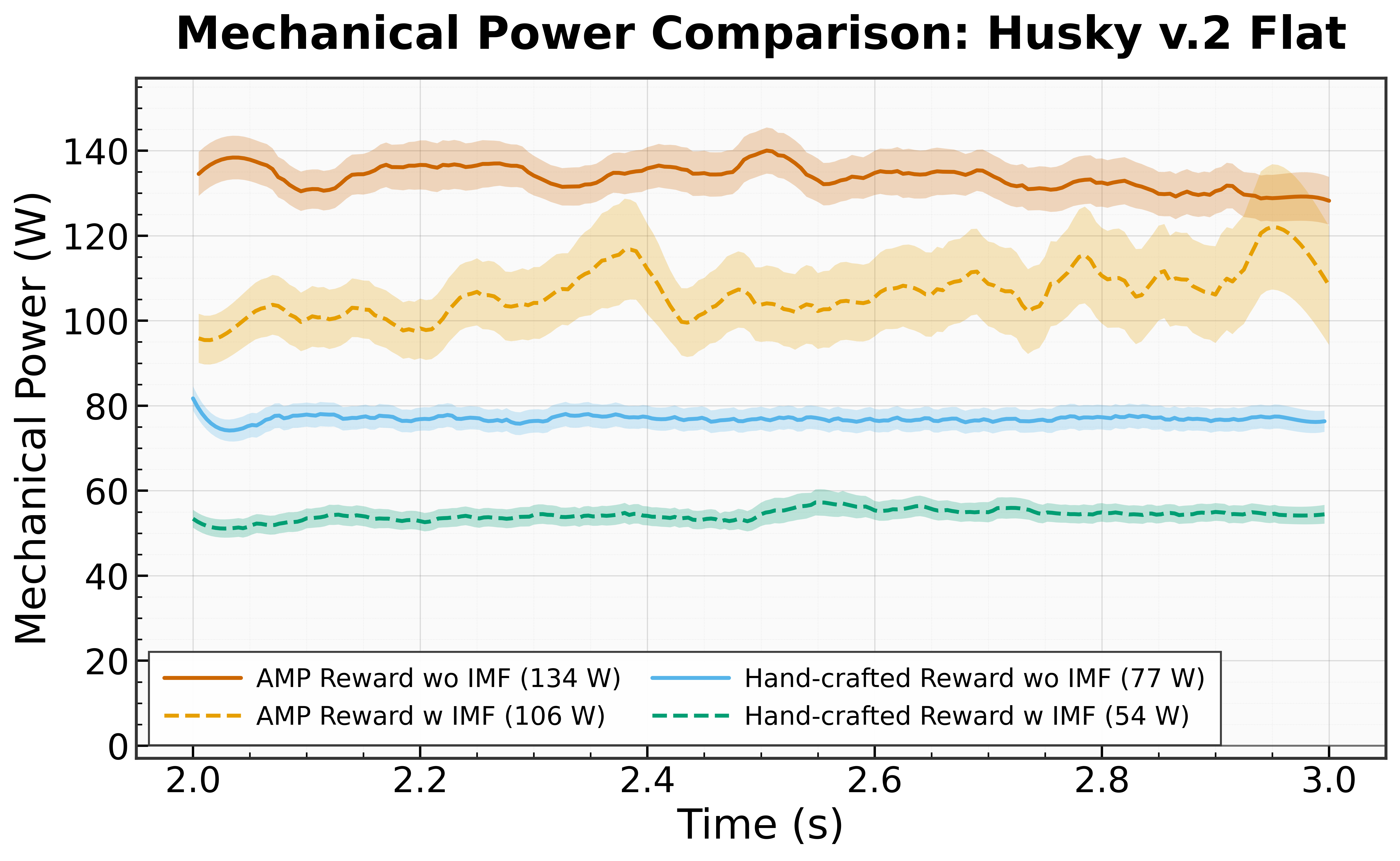}
    \caption{Comparison of average mechanical power across 100 flat-ground environments using four policies, with velocities ranging from 1.5 m/s to 2 m/s.}
    \label{fig:result-mech-power}
\end{figure}
% \FloatBarrier
% \begin{figure}[!ht]
%     \centering
%     \includegraphics[width = \linewidth]{figs/knee_torque_comparison.png}
%     \caption{Comparison of average knee torque across 100 flat ground environments using four policies, with velocities ranging from 1.5 m/s to 2 m/s.}
%     \label{fig:result-knee-torque}
% \end{figure}
% \FloatBarrier

Figures \ref{fig:result-mech-power} demonstrate the efficacy of IMF-augmented policies in reducing energy consumption by improving passive compliance. As shown in Figure \ref{fig:result-mech-power}, policies leveraging the IMF exhibit significantly reduced mechanical power consumption (calculated as \(\tau\cdot\dot{q}\)) compared to their baseline counterparts. On flat terrain at velocities of 1.5–2 m/s, the AMP+IMF policy reduces peak mechanical power by 35\ (from 100 W to 65 W), while the handcrafted+IMF policy achieves a 28\% reduction (from 70 W to 50 W). This efficiency gain correlates with reduced joint loading, as shown in Figure \ref{fig:result-mech-power}. The handcrafted+IMF policy reduces cumulative joint torques by 25–32\% compared to its baseline (e.g., rear left knee: 33\%, front right knee: 31\%), while the AMP+IMF policy achieves an 18–25\% reduction.

\section{CONCLUSIONS}

This work demonstrates how incorporating a physics-informed metric, the Impact Mitigation Factor (IMF), can significantly enhance energy efficiency and motion fidelity in legged robots. By uniting IMF with either Adversarial Motion Priors (AMP) or handcrafted rewards, we ensure policies learn not only explicit reference motions but also the underlying passive dynamics that enable biological systems to absorb and redirect impact forces. The resulting policies exhibit marked reductions in both mechanical power consumption and joint torques, translating into Cost of Transport (CoT) improvements of up to 32\% compared to baseline methods. Future work will expand this concept to broader contexts, including real-world hardware experiments and more aggressive maneuvers, aiming to fully maximize the potential of the IMF reward.

\printbibliography

\end{document}